\documentclass[journal]{IEEEtran}
\usepackage{cite}
\usepackage{amsmath,amssymb,amsfonts}
\usepackage{url}
\usepackage{amsmath}
\usepackage{algorithmic}
\usepackage{graphicx}
\usepackage{textcomp}
\usepackage{algorithm}
\usepackage{booktabs}
\usepackage{tabularx}
\usepackage{multirow}
\usepackage{arydshln}
\usepackage{array}
\usepackage{ragged2e}
\usepackage{adjustbox}
\usepackage{hyperref}
\begin{document}
\title{Surgery-R1: Advancing Surgical-VQLA with Reasoning Multimodal Large Language Model via Reinforcement Learning}
\author{Pengfei Hao, Shuaibo Li, Hongqiu Wang, Zhizhuo Kou, Junhang Zhang, Guang Yang, and Lei Zhu 
\thanks{Pengfei Hao, Shuaibo Li, and Hongqiu Wang are with the Hong Kong University of Science and Technology (Guangzhou), China (e-mail: phao467@connect.hkust-gz.edu.cn, sli270@connect.hkust-gz.edu.cn, hwang007@connect.hkust-gz.edu.cn).}
\thanks{Zhizhuo Kou is with the Hong Kong University of Science and Technology, Hong Kong SAR (e-mail: alankou@ust.hk).}
\thanks{Junhang Zhang is with Department of Thoracic Surgery, the Seventh Affiliated Hospital, Sun Yat-sen University, China (email:zhangjh33@mail.sysu.edu.cn).}
\thanks{Guang Yang is with Bioengineering/Imperial-X, Imperial College London, UK (e-mail: g.yang@imperial.ac.uk).}
\thanks{Lei Zhu is with the ROAS Thrust, Hong Kong University of Science and Technology (Guangzhou), China, and Department of Electronic and Computer Engineering, Hong Kong SAR, China (e-mail: leizhu@ust.hk).}
\thanks{Corresponding authors: Lei Zhu.}
}
\maketitle

\begin{abstract}
In recent years, significant progress has been made in the field of surgical scene understanding, particularly in the task of Visual Question Localized-Answering in robotic surgery (Surgical-VQLA). However, existing Surgical-VQLA models lack deep reasoning capabilities and interpretability in surgical scenes, which limits their reliability and potential for development in clinical applications. To address this issue, inspired by the development of Reasoning Multimodal Large Language Models (MLLMs), we first build the Surgery-R1-54k dataset, including paired data for Visual-QA, Grounding-QA, and Chain-of-Thought (CoT). Then, we propose the first Reasoning MLLM for Surgical-VQLA (Surgery-R1). In our Surgery-R1, we design a two-stage fine-tuning mechanism to enable the basic MLLM with complex reasoning abilities by utilizing supervised fine-tuning (SFT) and reinforcement fine-tuning (RFT). Furthermore, for an efficient and high-quality rule-based reward system in our RFT, we design a Multimodal Coherence reward mechanism to mitigate positional illusions that may arise in surgical scenarios. Experiment results demonstrate that Surgery-R1 outperforms other existing state-of-the-art (SOTA) models in the Surgical-VQLA task and widely-used MLLMs, while also validating its reasoning capabilities and the effectiveness of our approach. The code and dataset will be organized in \url{https://github.com/FiFi-HAO467/Surgery-R1}.
\end{abstract}


\section{Introduction}
\label{sec:introduction}
\begin{figure*}
    \centering
    \includegraphics[width=1\linewidth]{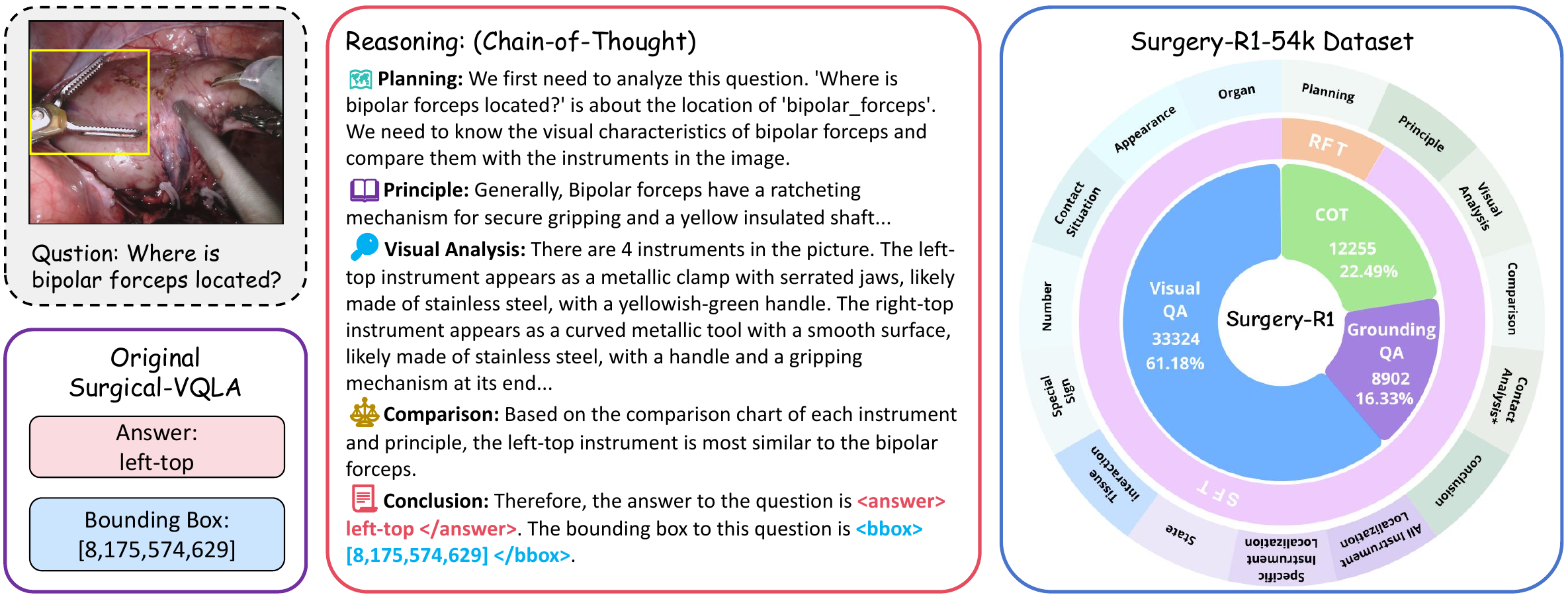}
    \caption{The left part displays the original Surgical-VQLA, where the input consists of surgical images and related questions, and the output includes answers along with their corresponding bounding boxes. The middle section illustrates the reasoning process of the questions, reflecting a complete Chain-of-Thought. The right part is the statistical chart of the Surgery-R1-54k dataset, which includes 12,255 complete Chains-of-Thought (CoTs), 33324 Visual Question-Answer pairs (Visual-QA pairs), and 8,902 Grounding Question-Answer pairs (Grounding-QA pairs).}
    \label{fig:1}
\end{figure*}

\IEEEPARstart{W}{ith} the rapid development of Robot-Assisted Surgery \cite{fiorini2022concepts}, surgical safety and precision have been significantly improved. Surgical scene understanding plays a vital role in this progress, serving as a key prerequisite for achieving intelligent surgery \cite{wang2024video,khan2025surgical}. In recent years, the rapid development of artificial intelligence has led to substantial progress in surgical scene understanding \cite{wang2023dynamic,khan2025surgical}. Among these advancements, the Visual Question Localized-Answering in Robotic Surgery (Surgical-VQLA) \cite{b3} task has garnered widespread attention as a new challenge in the field. Surgical-VQLA models are designed to answer fine-grained questions related to surgical instruments, target organs, and operational states, while simultaneously identifying the corresponding visual regions, as illustrated in Fig. \ref{fig:1} (left). This multimodal task provides a more comprehensive and detailed perspective for understanding surgical scenes \cite{b3,b6,hao2025enhancing}.

Recent Surgical-VQLA methods \cite{b3,b6,hao2025enhancing} have achieved notable performance improvements by employing pre-trained vision-language models as backbones in surgical scenes. Meanwhile, Large Language Models (LLMs) have achieved remarkable success in Natural Language Processing (NLP) \cite{openai2023gpt,touvron2023llama}. Building on this, researchers have introduced LLMs into the surgical domain \cite{b4} and further extended this direction by employing Multimodal Large Language Models (MLLMs) \cite{bai2023qwen,li2025otter} for Surgical-VQLA. These works \cite{wang2024surgical,wang2025endochat} constructed instruction-tuning datasets specifically tailored to surgical scenes to fine-tune MLLMs, leading to promising results. Collectively, these efforts demonstrate that instruction-fine-tuned MLLMs can improve task alignment and multimodal comprehension in surgical scenarios.

However, current Surgical-VQLA models lack explicit reasoning capabilities and interpretability, both of which are essential for safe and trustworthy deployment in surgical settings. These models typically produce direct answers without providing any insight into their underlying decision processes, thereby limiting the transparency and reliability of MLLMs in clinical applications. Given that MLLMs are prone to generating hallucinations \cite{liu2024survey}, simply presenting answers without justifications does not reveal the source of potential errors, thereby limiting further model optimization. Furthermore, evaluating model performance solely based on the final answer makes it challenging to determine whether the model has truly understood the surgical context or has merely guessed a plausible response. This lack of verifiability undermines surgeons' confidence in the model, as they cannot ascertain whether the model has correctly interpreted the complex visual information present in the surgical environment.

Recently, Reasoning LLMs have achieved significant advancements in performing deep reasoning when answering questions \cite{jaech2024openai,guo2025deepseek}. Notably, the success of DeepSeek-R1 \cite{guo2025deepseek} has demonstrated that emergent and robust reasoning capabilities can be imparted to LLMs through rule-based reinforcement learning (RL) \cite{zeng2024scaling} and Chain-of-Thought (CoT) \cite{zhang2022automatic}. Inspired by this, recent studies in have focused on extending RL training to Reasoning MLLMs \cite{peng2025lmm, zhan2025vision, feng2025video}. In medical domain, Reasoning MLLMs such as Med-R1 \cite{lai2025med} and HuatuoGPT-o1 \cite{chen2024huatuogpt} have utilized RL to enhance reasoning capabilities and interpretability in medical visual question answering (Medical-VQA), achieving notable success. However, in the surgical domain, the development of Reasoning MLLM for understanding surgical scenes remains an unexplored area.

In this paper, we introduce Surgery-R1, the first Reasoning MLLM for Surgical-VQLA. To equip MLLM with reasoning abilities in surgical scenes, we develop the Surgery-R1-54k dataset. This dataset extends the EndoVis-18-VQLA and EndoVis-17-VQLA datasets, including paired data
for CoT, Visual-QA, and Grounding-QA, as shown in Fig. \ref{fig:1} (right). We design specific reasoning processes for each type of question in Surgical-VQLA. For instance, in the case of instrument location questions, the reasoning process includes planning, principle, visual analysis, comparison, and conclusion, as illustrated in Fig. \ref{fig:1} (middle). The model first decomposes the reasoning steps according to the question type, then analyzes the general visual characteristics of the target instrument. Then, it describes each instrument in the image and finally determines which instrument most closely matches the visual characteristics of the target instrument through comparative analysis, thereby identifying the target instrument's location. This method of reasoning, based on visual comparative analysis, enables the model to achieve a deeper understanding of the surgical scene. In this process, Visual-QA serves as a subtask of CoT related to visual analysis of surgical images, while Grounding-QA involves the bounding boxes of organs and each instruments in the image. These three types of data collectively form the Surgery-R1-54k dataset.

To enable MLLMs to perform complex reasoning in surgical scenarios, we design a two-stage training method. In the first stage, we conduct supervised fine-tuning (SFT) \cite{dong2023abilities} of MLLMs using the data for SFT in Surgery-R1-54k dataset. This stage has two objectives: first, to fine-tune MLLMs for surgical scenes, as the original MLLMs are not trained on surgical datasets; second, to impart basic reasoning capabilities to MLLMs. In the second stage, we perform reinforcement fine-tuning (RFT) using a distinct portion of the Surgery-R1-54k dataset. Following DeepSeek-R1, we employ Group Relative Policy Optimization (GRPO) \cite{shao2024deepseekmath}, using rule-based rewards and group relative comparisons to stabilize training. These mechanisms not only reduce computational overhead but also encourage complex reasoning, making GRPO well-suited for surgical understanding tasks that require scalability and reliability.

Since Surgical-VQLA task requires generating bounding boxes along with answering questions, we design the Visual Grounding (VG) reward and Linguistic Answer (LA) reward for GRPO. Furthermore, through analyzing the reasoning process of Reasoning MLLM, we observe that the model sometimes exhibits positional hallucinations of surgical instruments. Specifically, the positional terms used during Reasoning MLLM reasoning sometimes do not align with the bounding boxes generated by the model. For example, while the model accurately localizes an instrument in the left-bottom, the described position might incorrectly be in the right-bottom. This suggests that although the model correctly identifies the instruments, it sometimes mislabels them during the reasoning process, ultimately leading to incorrect outcomes. Therefore, we additionally design the Multimodal Coherence (MC) reward to mitigate this hallucination and enhance the model's visual understanding capabilities. These reward mechanisms prevent our model from merely memorizing correct answers, as in SFT. Instead, our approach explores various solutions and optimizes towards outcomes defined by validated rewards, aiming to discover the most effective methods.

Overall, our contributions are as follows:
\begin{itemize}
\item We introduce Surgery-R1, the first Reasoning Multimodal Large Language Model for Surgical-VQLA. 
\item To enable more accurate answers and localization with reasoning, we design a two-stage training method. First, we perform supervised fine-tuning (SFT) on the basic MLLM, followed by reinforcement fine-tuning (RFT) using the Group Relative Policy Optimization (GRPO) algorithm. 
\item We design a rule-based reward system specifically tailored for the Surgical-VQLA task. Within this system, we design a Multimodal Coherence (MC) Reward mechanism to address and mitigate positional hallucination that may occur in surgical scenarios.
\item Extensive experiments demonstrate the effectiveness of Surgery-R1 on Surgical-VQLA task. Additionally, we develop the Surgery-R1-54k dataset, which includes paired data for Chain-of-Thought (CoT), Visual-QA, and Grounding-QA, along with corresponding surgical images.
\end{itemize}

\begin{figure*}[h]
    \centering
    \includegraphics[width=0.85\linewidth]{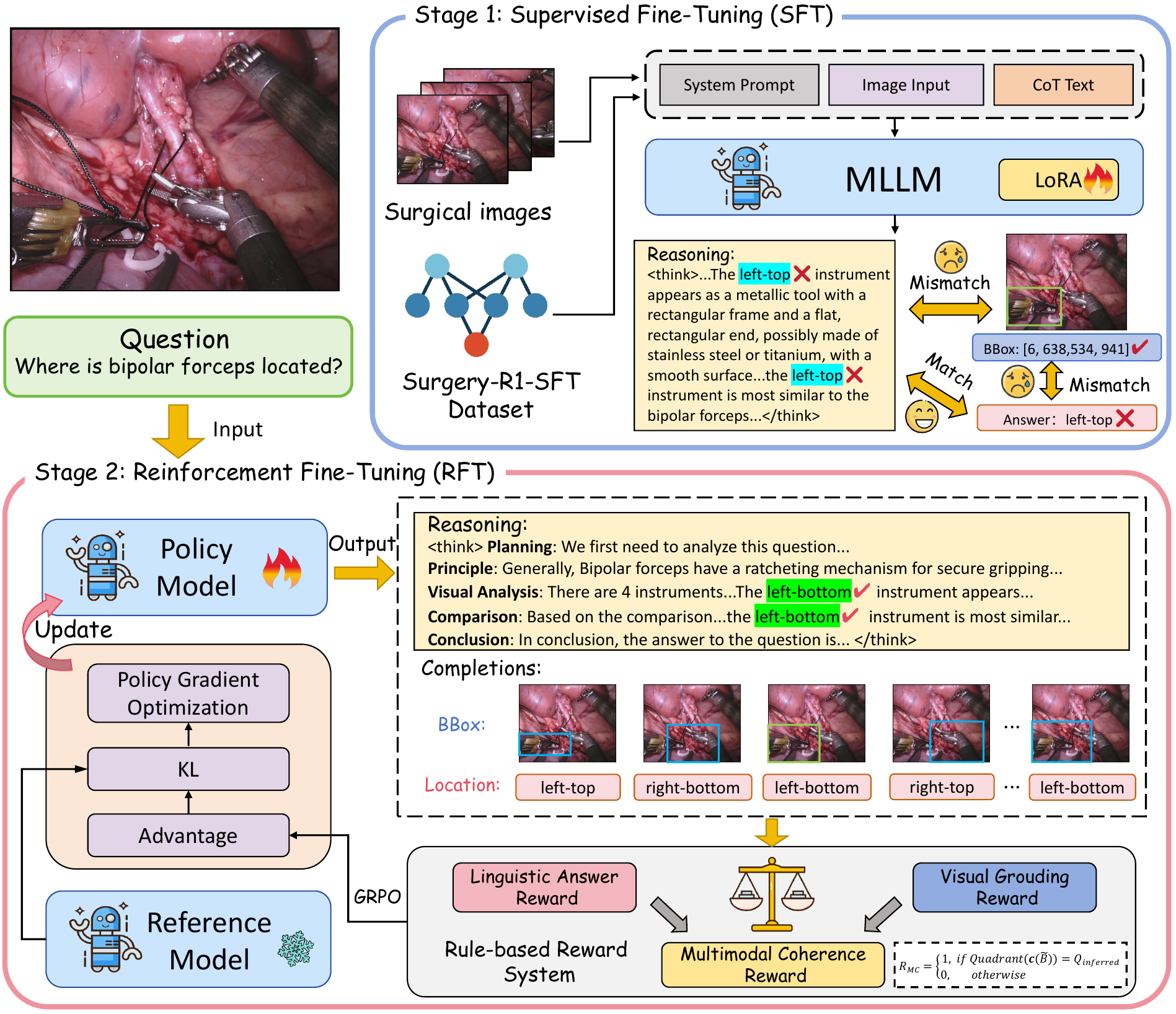}
    \caption{An overview of the framework of our Surgery-R1. Our training framework consists of two stages. First, we perform supervised fine-tuning (SFT) on the MLLM using the data for SFT in Surgery-R1-54k dataset (Surgery-R1-SFT subset). Next, we apply the Group Relative Policy Optimization (GRPO) algorithm for reinforcement fine-tuning (RFT). In the diagram, blue markers and $\times$ indicate incorrect text content, while green markers and $\checkmark$ indicate correct text content.}
    \label{fig:2}
\end{figure*}

\section{RELATED WORKS}

\subsection{Surgical-VQLA}

In recent years, Surgical-VQLA has gained widespread attention as a multimodal task for understanding surgical scenes. The objective of this task is to answer questions related to surgical instruments, the organs being operated on, and the operational status of these instruments based on input surgical images, while also providing corresponding bounding boxes. This requires the model to possess not only the ability to understand both visual and textual information but also precise localization capabilities to identify and mark key elements within the surgical scene.

The first Surgical-VQLA model, GVLE-LViT \cite{b3}, employs a pre-trained vision encoder and tokenizer to extract visual and textual features. These features are then integrated through gated fusion, with the Transformer serving as a decoder to derive the hidden features of the final layer. Finally, a classification head and a localization head are utilized to predict answers and bounding boxes. Subsequent research has largely built upon this encoder-decoder framework, introducing various improvements. For instance, CAT-ViL \cite{b6} designs a collaborative attention mechanism for feature fusion, enhancing the interaction between multimodal features. EnVR-LPKG \cite{hao2025enhancing} incorporates knowledge graphs generated by LLM, merging textual features of the knowledge graphs with surgical image features to strengthen the model's understanding of surgical scenes. With the advancement of MLLMs, new solutions have emerged for Surgical-VQLA. Surgical-MLLM \cite{wang2024surgical} introduced a novel fine-tuning instruction dataset and employed LoRA to fine-tune MLLMs, enabling them to comprehend surgical scenes. EndoChat \cite{wang2025endochat} proposed a larger and more comprehensive chat dataset for fine-tuning MLLMs, achieving state-of-the-art performance on the Surgical-VQLA task.

However, current Surgical-VQLA models fall short in deep reasoning and interpretability, which are crucial for surgical applications. These models typically provide direct answers without explaining the reasoning process, undermining the reliability of MLLMs in this domain. Although MLLMs enhance model performance in Surgical-VQLA, they can produce hallucinations, and merely providing answers does not reveal their origins, complicating optimization efforts. Furthermore, answers alone make it difficult to assess whether the model truly understands the surgical scene, as it might have simply guessed correctly. This limitation makes it challenging for surgeons to trust the model's judgments, as they cannot be certain whether it has accurately interpreted the complex information present in the surgical environment.

\begin{figure*}[h]
    \centering
    \includegraphics[width=0.85\linewidth]{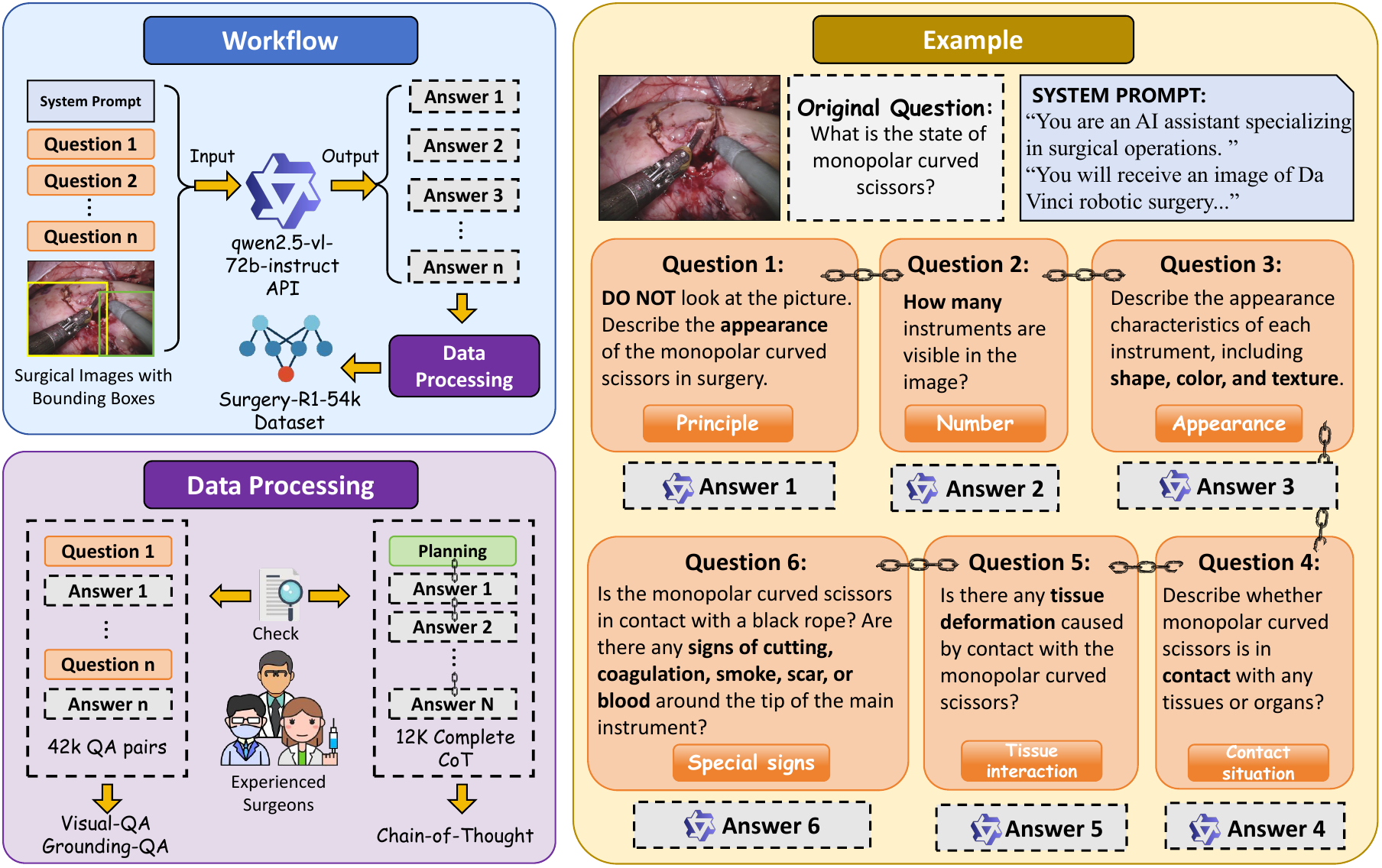}
    \caption{Details of data construction. The upper left part illustrates the workflow of data construction. The right part shows an example of constructing a chain of thought through sub-questions. The lower left part shows the details of data processing. We invited three experienced surgeons to review and revise our data, resulting in the final Surgery-R1-54k dataset.}

    \label{fig:3}
\end{figure*}

\subsection{Vision-Language Reinforcement Learning} 
Recently, as MLLMs have made significant progress in multimodal tasks, researchers have begun to explore how reinforcement learning can further enhance these models' performance and adaptability. Reinforcement learning offers MLLMs a dynamic adjustment mechanism, enabling them to better understand and respond to complex visual and language inputs. By incorporating Reinforcement Learning from Human Feedback (RLHF) \cite{sun2023aligning}, models can be optimized according to human preferences, reducing misunderstandings and hallucinations.

In this field, Direct Preference Optimization (DPO) \cite{rafailov2023direct} has emerged as a novel approach that simplifies the training process, allowing models to adjust directly based on annotated preference data. This method not only improves model alignment but also reduces reliance on resource-intensive manual annotations. Meanwhile, algorithms such as Proximal Policy Optimization (PPO) and Group-based Reinforcement Policy Optimization (GRPO) \cite{shao2024deepseekmath} offer new strategies for fine-tuning MLLMs. GRPO enhances model stability and computational efficiency by using group-based estimation and fixed-rule reward signals, making it particularly suitable for handling complex multimodal tasks.

In our research, we pioneered the application of GRPO to the multimodal understanding task within surgical scenarios, showcasing its potential in the surgical field. This work offers valuable insights and serves as a reference for future applications in other complex multimodal tasks within the surgical domain.

\section{METHODOLOGY}

\subsection{Overview of Surgery-R1 framework}
The framework Surgery-R1 is divided into two stages: Supervised Fine-Tuning (SFT) and Reinforcement Fine-Tuning (RFT), as illustrated in Fig. \ref{fig:2}.

In the first stage, SFT, we fine-tune the MLLM using the data for SFT in Surgery-R1-54k dataset (Surgery-R1-SFT subset) . This dataset includes surgical images, questions, complete CoT, answers, and corresponding bounding boxes. We design system prompts and input surgical images along with CoT text into the MLLM, enabling it to generate reasoning processes that assist in identifying key elements within surgical scenes. During this stage, the model adapts to surgical images through SFT, enhancing its understanding of surgical scenes and equipping it with basic reasoning capabilities.

However, SFT has some limitations. Firstly, SFT tends to memorize shortcuts for specific tasks rather than learning generalizable reasoning methods, which limits the MLLM's generalization and reasoning accuracy, preventing the model from fully understanding surgical scenes. Secondly, by analyzing the model's output reasoning process, we find that the MLLM has poor spatial orientation understanding of images, leading to hallucinations, as shown in Stage 1 of Fig. \ref{fig:2}, which has been concerned in related research \cite{cheng2024spatialrgpt,yang2024thinking, shi2025langloc}. Further analysis reveals potential inconsistencies between the bounding boxes generated by the MLLM and the spatial perception terms used in the reasoning process. Specifically, while the bounding box output by the MLLM is approximately accurate, the spatial descriptions in the reasoning process are incorrect. For example, in Stage 1 in Fig. \ref{fig:2}, the bounding box is located in the "left-bottom", but the reasoning process consistently describes it as "left-top".

To address these issues and further enhance the accuracy of responses and the precision of localization, we design the second stage, RFT, using the data for RFT in Surgery-R1-54k  dataset (Surgery-R1-RFT subset). In this stage, we employ the GRPO method, which optimizes the policy model by extracting a set of outputs for a question and calculating a set of relative advantages, rather than relying on a critic model. To adapt GRPO for the Surgical-VQLA task, we design a rule-based reward system, which includes MC Reward, LA Reward, and VG Reward. These rewards, calculated based on predefined rules, not only enhance the accuracy of the model's answers and grounding capabilities but also address inconsistencies in spatial orientation perception during the reasoning process.

\subsection{Dataset Construction}
To enable MLLMs to comprehensively understand surgical scenes and develop in-depth reasoning capabilities, we construct the Surgery-R1-54k dataset. For each category of questions in the EndoVis-18-VQLA and EndoVis-17-VQLA datasets, we design a detailed and complete reasoning process, known as Chain of Thought (CoT). Our CoT process guides the model to sequentially observe various visual details in surgical images and engage in step-by-step reasoning to arrive at the final answer. The reasoning process typically includes stages of planning, principle analysis, visual analysis, comparison, contact analysis (addressing unique issues related to the state of instruments), and conclusion, as illustrated in Fig. \ref{fig:1} (right).

Specifically, the planning stage involves breaking down the question into several thought steps for the model; during the principle analysis stage, the model analyzes the general visual characteristics of the target instrument; in the visual analysis stage, the model comprehensively observes the surgical image, including the number of instruments and the visual features of each instrument; during the comparison stage, the model determines which instruments in the image most closely match the visual characteristics identified in the principle analysis. For questions concerning the state of instruments, the contact analysis follows the comparison, where the model sequentially observes the contact between the target instrument and the tissue, checking for deformations in the tissue and the presence of special signs such as smoke or burn marks. By analyzing the contact details between the instrument and the tissue, the conclusion about the state of the instrument is reached.

In the CoT, the visual analysis and contact analysis consist of answers to multiple sub-questions, as shown in the Example in Fig. \ref{fig:3}. We use the API of powerful MLLM to answer these sub-questions, which are then revised by surgeons to produce the final answers. We compile these sub-question-answer pairs into Visual-QA data and sub-question-bounding box pairs into Grounding-QA data. Ultimately, the Surgery-R1-54k dataset includes paired Chain of Thought data, Visual-QA data, and Grounding-QA data.

The dataset construction process is illustrated in Fig. \ref{fig:3}, with the workflow in the upper left corner showing our data creation procedure. Initially, we meticulously design multiple sub-questions for each type of question in Surgical-VQLA. Subsequently, we design system prompts to input these sub-questions along with surgical images and bounding boxes into the qwen2.5-vl-72b-instruct model \cite{bai2025qwen2}, generating multiple corresponding sub-answers. Ultimately, the generated answers are compiled to form the Surgery-R1-54k dataset.

During the data processing process, we organize the generated answers into complete CoT following specific patterns. Additionally, we collected all sub-questions and their answers separately to form the Visual-QA data and Grounding-QA data, as shown in Fig. \ref{fig:3}. It is noteworthy that the reasoning processes within the CoTs are carefully crafted by experienced surgeons to ensure clinical relevance and logical rigor. Furthermore, to ensure data accuracy, all question-answer pairs and CoTs were reviewed and validated by 3 experienced surgeons. This rigorous review process guarantees the high quality of the dataset, providing a solid foundation for training MLLMs.

\subsection{Stage 1: Supervised Fine-Tuning}
The primary objective of the first stage of training is to fine-tune the MLLM using synthesized CoT data, Visual-QA data, and Grounding-QA data, thereby equipping the model with perceptual and deep reasoning abilities specific to surgical scenes, as shown in Fig. \ref{fig:2}. We design reasoning paths for the VQLA task to enhance the interaction between cognitive and perceptual processes. Specifically, we input images, questions, and CoTs or answers or bounding boxes from the Surgery-R1-SFT dataset into the MLLM for SFT, enabling the model to preliminarily understand surgical images and develop reasoning capabilities. This process integrates domain knowledge with human-like cognitive processes. We use the Qwen2.5-VL-3B model \cite{bai2025qwen2} as the base for SFT training, resulting in a model with foundational cognitive and visual perception abilities. This model demonstrates initial capabilities to integrate surgical domain knowledge and generate reasoning chains, laying a foundation for the second stage of reinforcement learning training. This approach bridges the gap between initial perception and expert-level understanding.

\subsection{Stage 2: Reinforcement Fine-Tuning}
\subsubsection{Group Relative Policy Optimization}
After completing the initial SFT training, we further enhanced the model's perception, localization, and reasoning abilities in surgical scenarios through reinforcement learning. As illustrated in Fig. \ref{fig:2}, we employed the Group Reinforcement Policy Optimization (GRPO) method. GRPO optimizes the policy model by generating a set of outputs and evaluating their relative advantages using group estimation. This method employs fixed rule-based reward signals to ensure consistency and stability of the rewards. During the policy update process, GRPO adjusts the model parameters by minimizing the KL divergence between the reference model and the current policy model. This adjustment encourages the generation of high-reward outputs, thereby enhancing the model's stability and computational efficiency in complex multimodal tasks \cite{shao2024deepseekmath}.

Specially, let \( q \) represent the question, and let \(\{ o_1, \ldots, o_G \}\) denote the set of sampled outputs. The previous policy model is \(\pi_{\theta_{\text{old}}}\). The optimization objective for the policy model \(\pi_\theta\) can be formally expressed as follows:

\begin{align}
\mathcal{J}_{GRPO}(\theta) &= \mathbb{E} \left[ q \sim P(Q), \{o_i\}_{i=1}^{G} \sim \pi_{\theta_{\text{old}}}(O \mid q) \right] \notag \\
&\quad \frac{1}{G} \sum_{i=1}^{G} \left( \frac{\pi_{\theta}(o_i \mid q)}{\pi_{\theta_{\text{old}}}(o_i \mid q)} A_i - \beta \mathcal{D}_{KL} \left( \pi_{\theta} \mid \mid \pi_{\text{ref}} \right) \right), \\
\mathcal{D}_{KL} (\pi_{\theta} \mid \mid \pi_{\text{ref}}) &= \frac{\pi_{\text{ref}}(o_i \mid q)}{\pi_{\theta}(o_i \mid q)} - \log \frac{\pi_{\text{ref}}(o_i \mid q)}{\pi_{\theta}(o_i \mid q)} - 1,
\end{align}
where \(\epsilon\) and \(\beta\) are hyperparameters, and \(A_i\) represents the advantage. \(\{r_1, \ldots, r_G\}\) are a set of rewards, and the outputs of each group are:

\begin{equation}
A_i = \frac{r_i - \text{mean}(\{r_1, \ldots, r_G\})}{\text{std}(\{r_1, \ldots, r_G\})}.
\end{equation}

\subsubsection{Reward Modeling}
We design a rule-based reward system specifically tailored for Surgical-VQLA, comprising 3 types of rewards: Linguistic Answer (LA) reward, Visual Grounding (VG) reward, and Multimodal Coherence (MC) reward.

\textbf{Visual Grounding Reward:} The VG reward improves the precision of model-generated bounding boxes. To this end, we utilize the Intersection Over Union (IoU) \cite{rezatofighi2019generalized} metric to measure the degree of spatial alignment between the predicted bounding boxes and their corresponding ground-truth annotations. Specifically, for a ground-truth bounding box \(B\) and a predicted bounding box \(\tilde{B}\), the VG reward is defined as follows:

\begin{equation}
R_{\text{VG}} = 
\begin{cases} 
\text{IoU}(B, \tilde{B}), & \text{if } \text{IoU}(B, \tilde{B}) \geq \tau \\ 
0, & \text{otherwise} 
\end{cases},
\end{equation}
where $\tau$ is the threshold that determines acceptable spatial grounding.

\textbf{Linguistic Answer Reward:} The purpose of the LA reward is to enhance the accuracy of the predicted answers. The LA reward evaluates the degree of match between the predicted answer and its corresponding ground-truth answer. Specifically, for a ground-truth answer \(A\) and a predicted answer \(\tilde{A}\), the LA reward is defined as follows:

\begin{equation}
R_{\text{LA}} = 
\begin{cases} 
1, & \text{if } A = \tilde{A} \\ 
0, & \text{otherwise} 
\end{cases}
\end{equation}

In this formula, the reward is 1 if the predicted answer exactly matches the ground-truth answer; otherwise, the reward is 0.

\textbf{Multimodal Coherence Reward:} 
In Section III.A, we discuss the issue of misalignment between the model's inferred location and the actual position of the bounding boxes, as illustrated in Stage 1 of Fig. \ref{fig:2}. Through extensive analysis of the inference process, we find that while the model often correctly identifies the approximate location of surgical instruments, the problem lies in the model's hallucination regarding spatial terms. In other words, the model correctly identifies the instrument but incorrectly describes its location. To mitigate this hallucination, we design a Multimodal Coherence Reward. We divide the image into four quadrants (left-top, left-bottom, right-top, right-bottom) and calculate the quadrant in which the center point of each predicted bounding box resides. We then match this result with the inferred location during reasoning; if they match, the reward is 1, otherwise, it is 0. Let the image dimensions be $W$ (width) and $H$ (height). For a predicted bounding box $\tilde{B}$ with center point coordinates $(x, y)$, the process can be expressed by the following equation:

\begin{equation}
\text{Quadrant}(x, y) =
\begin{cases}
\text{LT}, & \text{if } x < \frac{W}{2} \land y < \frac{H}{2}, \\
\text{RT}, & \text{if } x \geq \frac{W}{2} \land y < \frac{H}{2}, \\
\text{LB}, & \text{if } x < \frac{W}{2} \land y \geq \frac{H}{2}, \\
\text{RB}, & \text{if } x \geq \frac{W}{2} \land y \geq \frac{H}{2}.
\end{cases}
\end{equation}

\begin{equation}
R_{\text{MC}} =
\begin{cases}
1, & \text{if } \text{Quadrant}(\mathbf{c}(\tilde{B})) = Q_{\text{inferred}}, \\
0, & \text{otherwise}.
\end{cases}
\end{equation}

In the equations above, $\text{Quadrant}(x, y)$ determines which quadrant the center point $(x, y)$ of a predicted bounding box $\tilde{B}$ falls into, based on the image dimensions $W$ (width) and $H$ (height). The quadrants are labeled as $\text{LT}$ (left-top), $\text{RT}$ (right-top), $\text{LB}$ (left-bottom), and $\text{RB}$ (right-bottom). The Multimodal Coherence Reward $R_{\text{MC}}$ is 1 if the quadrant matches the inferred quadrant $Q_{\text{inferred}}$, and 0 otherwise. This reward helps ensure the model accurately aligns its predictions with spatial reasoning.

\begin{table*}[t]
\centering
\caption{Comparison experiments between our Surgery-R1 and other methods on EndoVis-18 and EndoVis-17 datasets}
\label{tab:1}
\resizebox{0.83\textwidth}{!}{%
\begin{tabular}{c|ccc|ccc}
\hline
\multirow{2}{*}{Models} & \multicolumn{3}{c|}{EndoVis - 18}          & \multicolumn{3}{c}{EndoVis - 17}                    \\ \cline{2-7} 
& Acc & F-score & mIoU & Acc & F-score & mIoU  \\

\hline 
\hline
VisualBERT \cite{b1}              & 0.6234          & 0.3413 & 0.7386          & 0.4048          & 0.3267          & 0.7136          \\
VisualBERT RM \cite{b1}           & 0.6341          & 0.3386 & 0.7379          & 0.4178          & 0.3346          & 0.7063          \\
MFH \cite{b41}                    & 0.5765          & 0.3265 & 0.7532          & 0.4255          & 0.3567          & 0.7237          \\
BlockTucker \cite{b42}            & 0.6192          & 0.3257 & 0.7521          & 0.4332          & 0.3533          & 0.7274          \\
MUTAN \cite{b43}                  & 0.5964          & 0.3524 & 0.7637          & 0.4365          & 0.3489          & 0.7130          \\
CAT-ViL DeiT \cite{b6} & 0.6476 & 0.3314 & 0.7746 & 0.4485 & 0.3528 & 0.7293      \\
LV-GPT \cite{seenivasan2023surgicalgpt}                 & 0.6775          & 0.4016 & 0.7592         & 0.4068          &  0.3626          & 0.6811          \\
GVLE-LViT \cite{b3} & 0.6647 & 0.3632 & 0.7641  & 0.4584 &0.2861 & 0.7258 \\ 
EnVR-LPKG \cite{hao2025enhancing} & 0.7071 & 0.4208 & 0.7936  & 0.4959 &0.4076 & 0.7441 \\
Surgical-MLLM \cite{wang2024surgical} & 0.6947 & 0.3325 & 0.8416  & 0.4068 &0.3412 & 0.7825 \\
EndoChat \cite{wang2025endochat} & 0.7147 & 0.4374 & 0.8689  & 0.5551 &0.2978 & \textbf{0.8662}\\
\hline
\hline
Surgery-R1 (our)   & \textbf{0.7356} & \textbf{0.4576} & \textbf{0.8721} & \textbf{0.5672} & \textbf{0.4422} & 0.8422 \\  \hline

\end{tabular}%
}
\end{table*}

\section{EXPERIMENTS}
\subsection{Dataset}
\subsubsection{Surgery-R1-54k Dataset}
The Surgery-R1-54k dataset is an extension of the EndoVis-18-VQLA and EndoVis-17-VQLA datasets. We generate the data using the qwen2.5-vl-72b-instruct API and invite experienced surgeons to review and modify it to ensure accuracy and professionalism. The Surgery-R1-54k dataset comprises 12,255 complete CoTs, 33342 Visual-QA pairs, and 8,902 Grounding-QA pairs. 

The CoT records the detailed reasoning process for each question in the Surgical-VQLA; Visual-QA provides additional questions distinct from Surgical-VQLA, covering aspects such as the number and appearance of instruments; Grounding-QA includes questions and their corresponding bounding boxes. 

Our test and training sets strictly adhere to the settings of EndoVis-18-VQLA and EndoVis-17-VQLA. In the division of the training set, 80\% of COTs, Visual-QA, Grounding-QA are used for the first stage of SFT, while the 20\% of CoTs are used for the second stage of RFT, as shown in Fig. \ref{fig:1} (right). This design ensures the effectiveness and stability of the model across different training stages.

\subsubsection{EndoVis-18-VQLA}
The EndoVis-18 Dataset, part of the MICCAI Endoscopic Vision Challenge 2018 \cite{b35}, contains video sequences from 14 robotic surgeries. Public annotations for EndoVis-18-VQLA\cite{b3} include question-answer pairs and bounding boxes related to the organ being operated, location of surgical instruments, and the state of instruments.

Following \cite{b3}, we use video sequences (1,5,16) as the test set, with the remainder for training. The training set has 1560 frames and 9014 question-answer pairs, while the validation set includes 447 frames and 2769 question-answer pairs.

\subsubsection{EndoVis-17-VQLA}
The EndoVis-17-VQLA Dataset, from the MICCAI Endoscopic Vision Challenge 2017\cite{b36}, includes 10 video sequences of robotic surgeries with publicly available annotations\cite{b3}.

It concludes 97 frames covering common organs, tools, and interactions, each with bounding boxes and VQA pairs. Following \cite{b3}, EndoVis-17-VQLA serves as an external validation dataset to test our model's generalization on unseen data, comprising 97 frames, 472 question-answer pairs, and 472 detection boxes, to evaluate model effectiveness and robustness in new surgical scenarios.

\subsection{Evaluation Metrics}
In evaluating the EndoVis-18-VQLA and EndoVis-17-VQLA datasets, we follow the methodology outlined in \cite{b3}, using Accuracy (Acc) \cite{b37}, F-score \cite{b37}, and mean Intersection over Union (mIoU) \cite{b38} to assess the model's accuracy in answer prediction and localization capabilities. These metrics help us comprehensively evaluate the model's performance in handling visual question answering tasks. 

\subsection{Implementation Details}
 We train our model using 8 NVIDIA GeForce RTX 4090 GPUs. Our model is implemented based on PyTorch. We initialize from Qwen2.5-VL-3B-Instruct \cite{bai2025qwen2}, employing per-GPU batch size 1 with 2-step gradient accumulation and bfloat16 mixed precision. In the first stage, we conduct SFT training for two epochs, setting the learning rate at 1e-6. Subsequently, in the second stage, we conduct RFT training for one epoch. The GRPO policy generates four candidate rationales per sample, with a sampling temperature of 0.7.  

\begin{table*}[t]
\centering
\caption{Comparison experiments between our Surgery-R1 and other MLLMs on EndoVis-18 and EndoVis-17 datasets}
\label{tab:2}
\resizebox{0.80\textwidth}{!}{%
\begin{tabular}{c|ccc|ccc}
\hline
\multirow{2}{*}{Models} & \multicolumn{3}{c|}{EndoVis - 18}          & \multicolumn{3}{c}{EndoVis - 17}                    \\ \cline{2-7} 
& Acc & F-score & mIoU & Acc & F-score & mIoU  \\
\hline 
\multicolumn{7}{c}{\textbf{Zero-shot MLLMs}}  \\  \hline
LLaVA-1.5 \cite{liu2023llava}    & 0.0462    & 0.0625 &   0.4436      &  0.0267         &   0.0334       &  0.3360     \\
LLaVA-Med \cite{li2023llava}      &  0.0549     & 0.0320   &  0.2648     & 0.0163          &  0.0122        &  0.3211     \\
Qwen2-VL \cite{wang2024qwen2}   & 0.0181      & 0.0147  &   0.1450     & 0.0254         &  0.0154         & 0.2043         \\
Qwen2.5-VL \cite{bai2025qwen2}         & 0.0322        & 0.0746  & 0.3144        & 0.0389          & 0.0349          & 0.2351       \\
\hline
\multicolumn{7}{c}{\textbf{Fine-tuned MLLMs}}  \\  \hline
Qwen2.5-VL (SFT) \cite{bai2025qwen2}              & 0.6499          & 0.3951 & 0.7325    & 0.4125          &  0.3144        &    0.7019      \\
Qwen2-VL (SFT) \cite{wang2024qwen2}           & 0.6210      & 0.3251 &   0.7416 
&  0.3962        &   0.2931       &  0.7336     \\
\hline \hline
Surgery-R1 (our)   & \textbf{0.7356} & \textbf{0.4576} & \textbf{0.8721}&\textbf{0.5672} & \textbf{0.4422} & \textbf{0.8422}  \\  \hline

\end{tabular}%
}
\end{table*}

\begin{table*}[t]
\centering
\caption{Ablation study on different variants of our approach on the EndoVis-18 and EndoVis-17 datasets. Baseline, M1, M2, and M3 represent diverse ablation models, while Surgery-R1 embodies our proposed comprehensive model, SFT represents Supervised Fine-Tuning, RFT represents Reinforcement Fine-Tuning and CoT presents Chain of Thought}
\label{tab:3}
\resizebox{0.85\textwidth}{!}{%
\begin{tabular}{@{}c|c|c|c|ccc|ccc@{}}
\toprule
\multirow{2}{*}{Models} & \multirow{2}{*}{SFT}  & \multirow{2}{*}{RFT}      & \multirow{2}{*}{CoT} & \multicolumn{3}{c|}{EndosVis-18}   & \multicolumn{3}{c}{EndosVis-17}                     \\ \cline{5-10} 
     & &    &           & Acc             & F-score         & mIoU            & Acc             & F-score         & mIoU            \\ \hline
Baseline       &    &  & & 0.0322        & 0.0746  & 0.3144         & 0.0389          & 0.0349          & 0.2351           \\
M1      & \checkmark  &  &  & 0.6499          & 0.3951 & 0.7325          & 0.4125          &  0.3144        &    0.7210        \\
M2      & \checkmark  &  & \checkmark  & 0.6627   & 0.3723 & 0.7526          & 0.4021          &  0.3550       &    0.7263        \\
M3    &\checkmark     & \checkmark  & & 0.7297          & 0.3921          & \textbf{0.8734}          & 0.5130          & \textbf{0.4523}          & 0.8024 \\

\hline
\vspace{-0.6mm}
Surgery-R1    & \checkmark    & \checkmark  & \checkmark & \textbf{0.7356} & \textbf{0.4576} & 0.8721 & \textbf{0.5672} & 0.4422 & \textbf{0.8422 }         \\
\bottomrule
\end{tabular}%
}
\end{table*}
\subsection{Ablation Study Setting}
To validate the effectiveness of the reasoning capabilities in our proposed Surgery-R1 method and the role of each training step, we designed a series of ablation experiments. The experimental conditions are as follows: (1) Baseline: We use the original Qwen2.5-VL model as the baseline, which has not been trained on surgical data. (2) M1: We perform supervised fine-tuning (SFT) on the baseline model using the Surgery-R1 dataset without Chains of Thought (CoT) to explore the impact of supervised fine-tuning. (3) M2: We perform SFT on the baseline model using the Surgery-R1 dataset with CoT to evaluate the effectiveness of reasoning capabilities. 
(4) M3: We first perform SFT on the baseline model using the Surgery-R1 dataset without CoT. Subsequently, we conducted RFT training to validate the effectiveness of RFT in enhancing model performance.

\subsection{Results}

\subsubsection{Comparisons With the State-of-the-Art Approaches}

In this section, we compare our proposed Surgery-R1 method with the current state-of-the-art Surgical-VQLA model to evaluate its performance on the EndoVis-18 and EndoVis-17 datasets. Table \ref{tab:1} presents the performance of each model in terms of Acc, F-score, and mIoU.

On the EndoVis-18 dataset, the Surgery-R1 method demonstrates superior performance across all metrics, achieving an accuracy of 0.7356, an F-score of 0.4576, and an mIoU of 0.8721. These results significantly outperform those of other Surgical-VQLA models. Specifically, compared to the state-of-the-art EndoChat model, our model shows improvements in all three metrics. Surgery-R1 achieves an accuracy that is 2.09 percentage points higher than EndoChat's 0.7147; an F-score that is 2.02 percentage points higher than EndoChat's 0.4374; and an mIoU that is 0.32 percentage points higher than EndoChat's 0.8689. These findings indicate that Surgery-R1 has a significant advantage in overall performance.

\begin{figure*}[h]
    \centering
\includegraphics[width=0.95\linewidth]{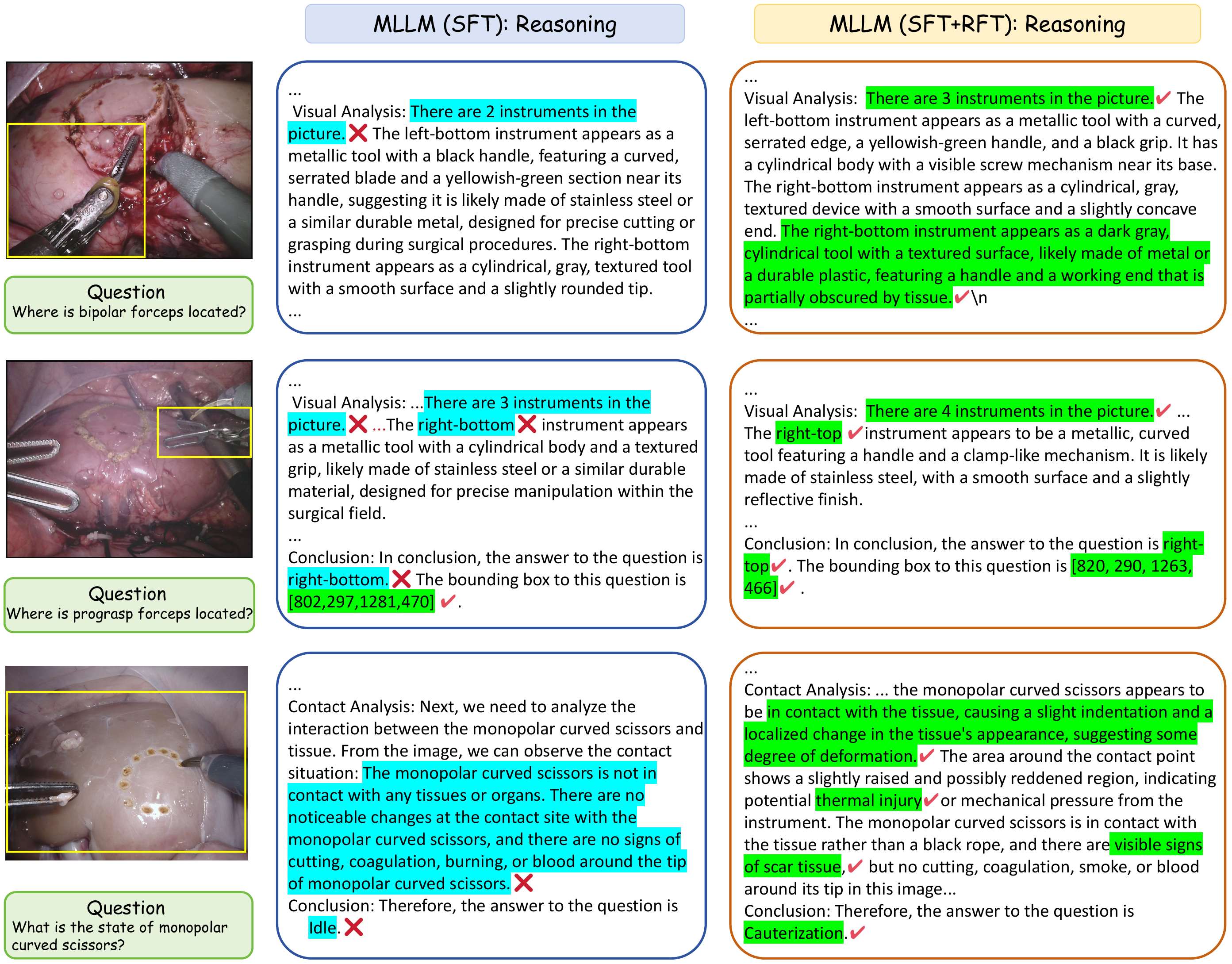}
    \caption{Qualitative analysis of the reasoning process of MLLM after SFT and after SFT+RFT. The figure shows significant improvements in spatial positional illusions after RFT. Blue markings and \(\times\) indicate incorrect text content, while green markings and \(\checkmark\) indicate correct text content.}
    \label{fig:4}
\end{figure*}

On the EndoVis-17 dataset, Surgery-R1 also exhibits outstanding performance, with an accuracy of 0.5672, an F-score of 0.4422, and an mIoU of 0.8422. Although the mIoU is slightly lower than EndoChat, Surgery-R1 surpasses EndoChat by 1.21 percentage points in accuracy and by 3.46 percentage points in F-score, demonstrating more prominent results. In summary, the Surgery-R1 method exhibits leading performance on both datasets, demonstrating its effectiveness and advantages in surgical visual question answering tasks.

\subsubsection{Comparison to zero-shot and SFT evaluations with other
MLLMs}
We compare the Surgery-R1 method with zero-shot and fine-tuned visual language models (VLMs) to evaluate its performance on the EndoVis-18 and EndoVis-17 datasets, as shown in Table \ref{tab:2}.

In the zero-shot evaluation, LLaVA-1.5, LLaVA-Med, Qwen2-VL, and Qwen2.5-VL all performed poorly on both datasets, with accuracy and F-score below 0.05 and mIoU not exceeding 0.45, significantly lower than Surgery-R1. This indicates the limited applicability of zero-shot VLMs in the Surgical-VQLA task.

In the fine-tuning evaluation, Qwen2.5-VL (SFT) and Qwen2-VL (SFT) achieved maximum accuracies of 0.6499 and 0.6210, respectively, on the EndoVis-18 dataset, with mIoU values of 0.7325 and 0.7416. On the EndoVis-17 dataset, Qwen2.5-VL (SFT) achieved an mIoU of 0.7019, while Qwen2-VL (SFT) achieved an mIoU of 0.7336. Although fine-tuning improved performance, it still did not reach the level of Surgery-R1. In contrast, Surgery-R1 demonstrated superior performance on both datasets, proving its effectiveness and advantages in the Surgical-VQLA task.

\subsubsection{Effects of SFT}

To validate the performance improvement brought by fine-tuning (SFT), we conducted a comparative analysis between the M2 model and the Baseline model. The M2 model underwent SFT on the Surgery-R1 dataset without CoT data. This comparison allows us to assess the impact of SFT on model performance. As detailed in Table \ref{tab:3}, M2 shows significant enhancements in accuracy, F-score, and mIoU across both datasets. On the EndoVis-18 dataset, M2 achieves improvements of 63.05\% in accuracy, 29.75\% in F-score, and 43.82\% in mIoU. Similarly, on the EndoVis-17 dataset, M2 demonstrates increases of 36.32\% in accuracy, 32.01\% in F-score, and 49.12\% in mIoU. These findings suggest that SFT significantly enhances the model's ability to comprehend surgical scenes, thereby improving its performance in complex visual question answering tasks.

\subsubsection{Effects of RFT}
To validate the performance improvement brought by RFT, we conducted a comparative analysis between the M3 model and the M1 model. The M3 model was subjected to RFT based on the M1 model. This comparison allows us to assess the impact of RFT on model performance. As detailed in Table \ref{tab:3}, the M3 model exhibits significant improvements in accuracy, F-score, and mIoU across both the EndoVis-18 and EndoVis-17 datasets. Specifically, on the EndoVis-18 dataset, the M3 model's accuracy and mIoU are higher than those of M1 by 7.98\% and 14.09\%, respectively, although the F-score is slightly lower by 0.3\%. Similarly, on the EndoVis-17 dataset, the M3 model's accuracy, F-score, and mIoU surpass those of M1 by 10.05\%, 13.79\%, and 8.14\%, respectively. These results indicate that RFT significantly enhances model performance, particularly in terms of accuracy and mIoU, despite a slight decrease in F-score in some cases. This demonstrates the effectiveness of RFT in optimizing the model's ability to understand complex surgical scenes.

\subsubsection{Effects of Reasoning}

To validate the impact of reasoning capabilities on model performance, we conduct a comparative analysis between the M2 model and the M1 model. The M2 model was fine-tuned using CoT data, whereas the M1 model was not trained on CoT data. This comparison allows us to assess the influence of reasoning capabilities on model performance. As detailed in Table \ref{tab:3}, the M2 model shows improvements across various metrics on both the EndoVis-18 and EndoVis-17 datasets. Specifically, on the EndoVis-18 dataset, the M2 model's accuracy and mIoU are higher than those of M1 by 1.26\% and 2.01\%, respectively, although the F-score is slightly lower by 2.28\%. On the EndoVis-17 dataset, the M2 model's F-score and mIoU exceed those of M1 by 4.06\% and 0.53\%, respectively, but the accuracy is slightly lower by 1.04\%.

Additionally, we compared the Surgery-R1 model with the M3 model. The Surgery-R1 model was fine-tuned with CoT data, while the M3 model was not. The results indicate that the Surgery-R1 model's accuracy and F-score on the EndoVis-18 dataset are higher than those of M3 by 0.59\% and 6.55\%, respectively, although the mIoU is slightly lower by 0.13\%. On the EndoVis-17 dataset, the Surgery-R1 model's accuracy and mIoU surpass those of M3 by 5.42\% and 3.98\%, respectively, but the F-score is slightly lower by 1.01\%.

These results suggest that reasoning capabilities can significantly enhance the overall performance of the model. By incorporating reasoning abilities, the model's understanding and processing capabilities in complex surgical scenes are improved.

\subsubsection{Qualitative Analysis of Reasoning Quality}

Through qualitative analysis of reasoning quality, we observe that models enhanced with RFT exhibit significant improvements in reasoning quality compared to those subjected only to SFT, as illustrated in Fig. \ref{fig:4}. Analyzing the reasoning process of model only trained with SFT, we find that it frequently makes errors in identifying the number of surgical instruments in images. Additionally, the model often displays mismatches between bounding boxes and textual descriptions and experiences severe hallucinations in understanding locative terms. For instance, in example 1, the model incorrectly identifies two instruments when there are actually three. Although the bounding boxes correctly mark the positions of the instruments, the textual descriptions are inaccurate.

In contrast, model enhanced with RFT demonstrates higher reasoning accuracy. The RFT-enhanced model is able to correctly identify the number of instruments and more accurately align bounding boxes with textual descriptions. This improvement is evident in example 2 where the RFT model correctly identify three instruments and match the bounding box with the description of its position. Consequently, the issue of hallucination, where bounding boxes and textual descriptions do not match, is significantly reduced.

These findings indicate that RFT not only improves the accuracy of instrument count recognition but also enhances the coherence between visual outputs and textual descriptions. This enhancement significantly boosts the model's reasoning capabilities, enabling it to provide more reliable and precise information in complex surgical environments.


\section{CONCLUSION}
In this paper, we introduce Surgery-R1, the first Reasoning MLLM for Surgical-VQLA that possesses deep reasoning capabilities and interpretability. By developing the Surgery-R1-54k dataset, designing an innovative training framework, and implementing a reward system, we successfully endow the MLLM with complex reasoning abilities in surgical scenarios. Experimental results demonstrate that Surgery-R1 outperforms existing state-of-the-art models in the Surgical-VQLA task and widely-used MLLMs, validating its reasoning capabilities and the effectiveness of our approach. Our research highlights the significant potential of reasoning models in surgical scene understanding, laying a foundation for the advancement of intelligent surgery. We anticipate that future research will further expand the capabilities of Reasoning MLLM, providing more support and possibilities to enhance surgical safety and precision.

\bibliographystyle{IEEEtran}
\bibliography{ref}
\end{document}